\documentclass{article}
\usepackage{spconf,amsmath,epsfig}
\usepackage{multirow}
\usepackage{multicol}
\usepackage{amsfonts}
\usepackage{float}
\usepackage{stfloats}
\usepackage{color}
\let\OLDthebibliography\thebibliography
\renewcommand\thebibliography[1]{
  \OLDthebibliography{#1}
  \setlength{\parskip}{0pt}
  \setlength{\itemsep}{0pt plus 0.3ex}
}

\usepackage{booktabs}
\usepackage{url}

\begin{document}\sloppy
\newcommand{\suh}[1]{\textcolor{blue}{#1}}
% Example definitions.
% --------------------
\def\x{{\mathbf x}}
\def\L{{\cal L}}

% Title.
% ------
\title{SASFormer: Transformers for Sparsely Annotated Semantic Segmentation}
%
% Single address.
% ---------------
\name{Hui Su$^{\dagger}$, Yue Ye$^{\dagger}$, Wei Hua$^{\dagger}$, Lechao Cheng$^{\dagger}$\sthanks{Corresponding author.}, Mingli Song$^{\dagger\dagger}$}

%Address and e-mail should NOT be added in the submission paper. They should be present only in the camera ready paper. 
\address{\{suhui, yeyue, huawei, chenglc\}@zhejianglab.com, brooksong@zju.edu.cn \\
        $^{\dagger}$ Zhejiang Lab\\ $^{\dagger\dagger }$ Zhejiang University}

\maketitle

\begin{abstract}
Semantic segmentation based on sparse annotation has advanced in recent years. It labels only part of each object in the image, leaving the remainder unlabeled. 
%Existing approaches based on the pseudo-label are time-consuming and often necessitate a multi-stage training strategy. 
Most of the existing approaches are time-consuming and often necessitate a multi-stage training strategy.
In this work, we propose a simple yet effective sparse annotated semantic segmentation framework based on segformer, dubbed SASFormer, that achieves remarkable performance. Specifically, the framework first generates hierarchical patch attention maps, which are then multiplied by the network predictions to produce correlated regions separated by valid labels. Besides, we also introduce the affinity loss to ensure consistency between the features of correlation results and network predictions. Extensive experiments showcase that our proposed approach is superior to existing methods and achieves cutting-edge performance. The source code is available at \url{https://github.com/su-hui-zz/SASFormer}.
% The abstract should appear at the top of the left-hand column of text, about 0.5 inch (12 mm) below the title area and no more than 3.125 inches (80 mm) in length.  Leave a 0.5 inch (12 mm) space between the end of the abstract and the beginning of the main text.  The abstract should contain about 100 to 150 words, and should be identical to the abstract text submitted electronically along with the paper cover sheet.  All manuscripts must be in English, printed in black ink.
\end{abstract}
\section{Introduction}
% 语义分割的问题
% Semantic segmentation is a fundamental task in computer vision, which aims to label each pixel in an image. Although semantic segmentation has witnessed continuous improvements in recent years, effect of semantic segmentation needs a lot of time and manpower to make pixel by pixel annotation.
% Due to the high cost of labeling, researchers have to explore how to maintain good segmentation effect while only making partial labeling. 
% Sparsely annotated semantic segmentation (SASS) comes into being, which provides sparse annotations for each object in an image\cite{liang2022tree}, such as point-wise\cite{bearman2016s}\cite{qian2019weakly} and scribble-wise\cite{lin2016scribblesup}\cite{xu2021scribble}.

Semantic segmentation is an essential problem in computer vision, which seeks to identify each pixel in an image. Although semantic segmentation has observed ongoing improvements in recent years, impact of semantic segmentation demands a lot of time and labor to conduct pixel by pixel annotation. Due to the high cost of labeling, researchers have to examine techniques to maintain adequate segmentation performance while only making partial labeling. Sparsely annotated semantic segmentation (SASS) comes into existence, which provides sparse annotations for each object in an image~\cite{liang2022tree}, such as point-wise~\cite{bearman2016s,qian2019weakly} and scribble-wise~\cite{lin2016scribblesup,xu2021scribble} supervision.

%, block-wise ones\cite{liang2022tree}.

% 稀疏标注语义分割是什么，有什么缺陷
% Sparse annotation semantic segmentation belongs to a family of weakly supervised semantic segmentation (WSSS). 
% It only labels part region of each object in the image, leaving the remainder unlabeled.
% Due to the lack of overall object range and precise object boundary, performance of segmentation is hampered.
% It is necessary to introduce more prior knowledge in order to estimate unlabeled pixels more accurately. 

\begin{figure}[h]
\centering
\includegraphics[width=0.46\textwidth,height=0.38\textwidth]{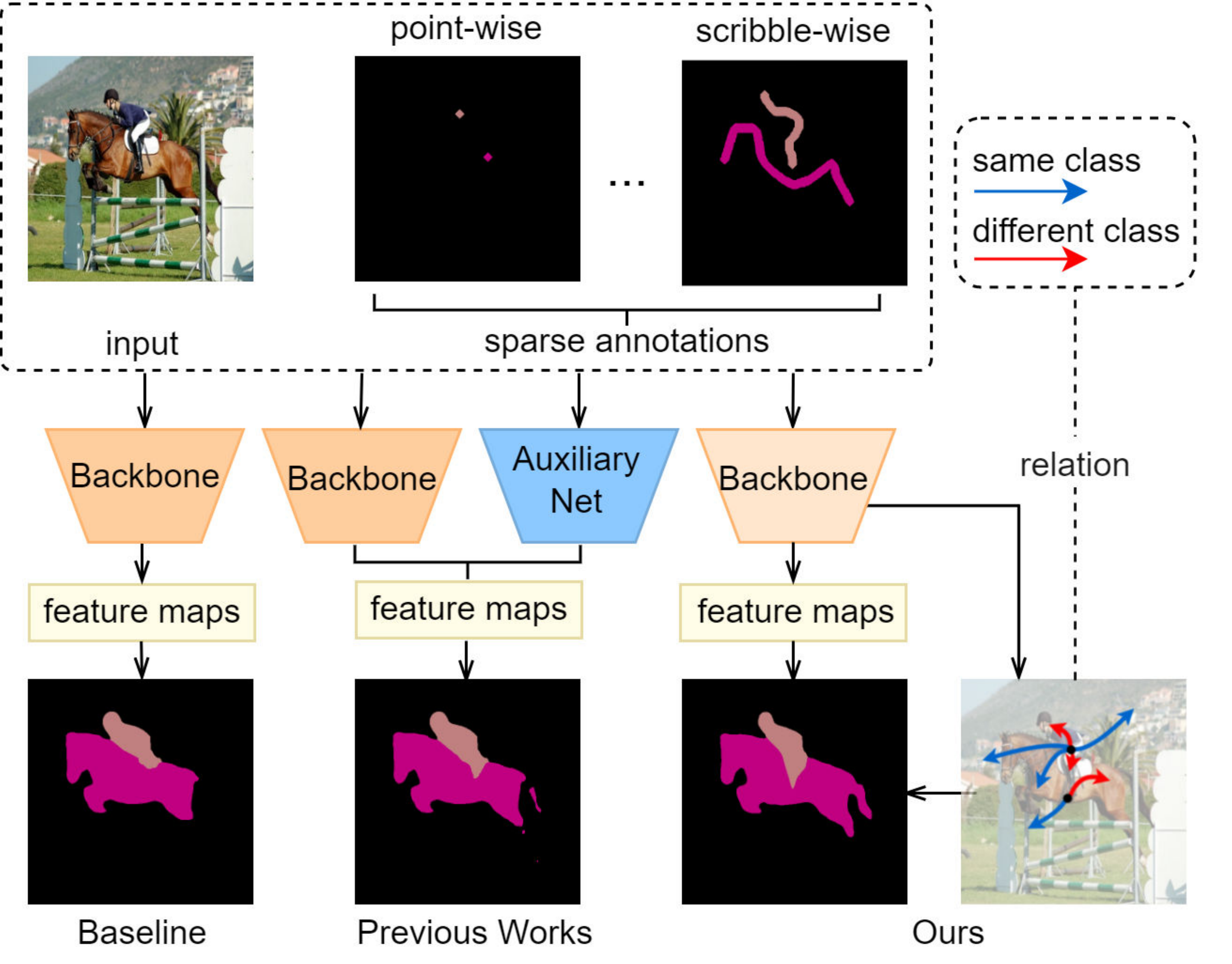} %[width=8\linewidth]
%[width=0.92\textwidth,height=0.36\linewidth]
\caption{Semantic segmentation with sparse annotation. The baseline trained only with sparse annotations is incapable of recognizing the full object. In the majority of earlier efforts, auxiliary networks containing information about other tasks or previous stage. Our SASFormer utilizes inherent global dependencies of the transformer to achieve state-of-the-art performance without developing a complicated framework.} \label{fig:teaser} %  % 大图名称 clip, trim=0.3cm 0.4cm 0.3cm 6.2cm,
  %图片引用标记
\vspace{-2mm}
\end{figure}

Sparse annotation semantic segmentation is a kind of weakly supervised semantic segmentation (WSSS)~\cite{zhang2021weakly}. It marks just a piece of each visual item, leaving the rest unidentified. Due to the absence of a precise object boundary, segmentation performance is substantially impaired. To more precisely estimate unlabeled pixels, it is required to discover more hidden information.

% 各类方法的不足
Existing methodologies for SASS can be categorized as regularization loss, multitask auxiliary, consistent learning, and the pseudo-label method. Regularization losses~\cite{tang2018regularized,marin2019beyond} utilize MRF/CRF potentials to implement clustering of low-dimensional information. Multitask assistance~\cite{wang2019boundary} increases segmentation performance by adding an auxiliary counterpart, for example boundary detection. Consistency learning~\cite{chen2021seminar,ke2021universal} introduces multiple networks to compensate for insufficient information provided in sparse annotations. The pseudo-label approach typically involves a multi-stage training procedure.
Most of the above methods necessitate the use of additional complex frameworks to facilitate training, resulting in a complicated and time-consuming training process.

%上述的方法多数需要额外的复杂framwork辅助训练，训练过程复杂耗时。 近年来，vision transformers\cite{zheng2021rethinking,xie2021segformer} 在分割领域逐渐发展起来。Vision transformer自带的 self-attention mechanism can capture more effective relationships between pixels。这很有利于信息从labeled pixels 向。。的传播，挖掘 unlabeled pixels的潜藏信息。我们利用vision transformr探索单阶段有效的稀疏标注分割方式，旨在利用简单的方式实现最优的效果。
In recent years, vision transformers~\cite{xie2021segformer} have steadily evolved in various fields~\cite{fang2022cross, xue2022protopformer}. The self-attention mechanism in the vision transformer can capture effective relationships between pixels, which is beneficial for information propagation from labeled pixels to unlabeled pixels. Inspired by this, We investigate single-stage effective sparsely annotated semantic segmentation with vision transformer, aiming to achieve the state-of-the-art results in a straightforward manner, as shown in Fig~\ref{fig:teaser}.
In this work, we propose SASFormer, a simple yet effective framework based on segformer\cite{xie2021segformer}, as the first effort at sparsely annotated semantic segmentation with Vision Transformer. With the inherent patch-to-patch attention of cascaded transformer blocks, we first create patch attention maps at multiple levels. The patch-level attention maps are then multiplied by network predictions in order to correlate labeled and unlabeled image regions. Following this, an affinity loss function is applied to assure the consistency of the correlation statistics and network predictions. We finally consider embedding priors from unlabeled areas to boost the performance by combining affinity loss and conventional segmentation loss. Our contributions can be summarized in the following manner:
\begin{itemize}
    \item We propose SASFormer as the first trustworthy baseline for SASS with a visual transformer to model relationships between distinct areas and offer category recommendations to unlabeled regions.
    \item We provide a novel affinity loss function to assure the similarity prior in SASS: areas of identical objects share similarity in both low-dimensional and high-dimensional feature space.
    \item The proposed SASFormer manifests remarkable performances in both point- and scribble-annotated SASS assignments on PASCAL VOC. 
\end{itemize}

\section{Related Works}
\subsection{Sparsely Annotated Semantic Segmentation}
Sparsely annotated semantic segmentation aims to address the segmentation problem using minimally labeled visual areas. What's the point\cite{bearman2016s} brought point-level labels to the semantic segmentation problem for the first time, which boost the performance by introducing objectiveness priors.
After that, ScribbleSup\cite{lin2016scribblesup} extends point-level labels into the scribbling ones in order to bridge the performance gap between fully annotated methods and unlabeled ones. ScribeSup constructs a graphical model to transmit information from scribbles to unknown pixels, hence providing semantic guidance for unknown pixels. Since then, approaches for sparse annotated semantic segmentation have emerged and developed. In an attempt to enhance performance at the lowest feasible cost, researchers have begun investigating the association between labeled and unlabeled regions. For example, Tang et al.~\cite{tang2018regularized} offered a variety of regularized losses for sparsely supervised segmentation based on dense CRF and kernel cut. BPG~\cite{wang2019boundary} tries to integrate semantic characteristics and textural information by constructing a prediction refinement network, whilst a boundary regression network is introduced to facilitate performance by generating clearly defined semantically separate parts. Recently, SPML~\cite{ke2021universal} tackles this challenge by developing a semi-supervised metric learning approach with four unique forms of attraction and repulsion relationships
TEL~\cite{liang2022tree} presents a tree energy loss, in which minimal spanning trees are constructed to represent low-level and high-level pair-wise affinities. The majority of these techniques either use multi-stage processes for progressive inference or include intricate frameworks for supplementary training. While in this work, we make the first attempt to model relationships between distinct areas and offer category recommendations to unlabeled regions based on vision transformer.

\subsection{Transformer}
The Transformer was initially proposed to model long-term dependencies in natural language processing tasks~\cite{vaswani2017attention}. In 2020, Alexey Dosovitskiy introduced the pure Transformer architecture, which achieved remarkable results in image classification~\cite{dosovitskiy2020image}. Subsequently, the Transformer has been widely employed in various computer vision tasks, such as object detection~\cite{carion2020end,su2022re, qiu2023team}, semantic segmentation~\cite{xie2021segformer, li2023boosting} and video processing~\cite{zhang2021token, zhang2022long}. Recently, researchers have started to utilize the long-term dependencies mechanism of the Transformer to capture the relationship between image categories and local image features, aiming to optimize weakly supervised problems~\cite{gao2021ts}. However, these methods mainly focus on utilizing the correlation between class tokens and patch tokens, while few works consider the optimization and utilization of the correlation among different patch tokens. Our work takes an early attempt to delve deeply into the issue of imprecise correlation among different patch tokens in the Transformer block and proposes a simple yet effective solution.
% \suh{\textbf{Transformer: }}
% \suh{
% Transformer is originally proposed to capture long-term dependencies in natural language processing tasks~\cite{vaswani2017attention}. In 2020, Alexey introduced the pure transformer architecture and achieved remarkable results in image classification~\cite{dosovitskiy2020image}. Since then, transformer has been widely used in many computer vision tasks, such as object detection~\cite{carion2020end,su2022re} and semantic segmentation~\cite{xie2021segformer}. Recently, researchers start to use the long-term dependencies mechanism of the transformer to capture the relationship between image categories and local image features in order to optimize weakly supervised problems~\cite{gao2021ts}. However, These methods focus on utilizing the correlation between class token and patch tokens, while few tasks take into account the utilization and optimization of the correlation among patch tokens. Our work makes an early attempt to delve deeply into the issue of imprecise correlation among different patch tokens in the transformer block and puts forth a simple yet effective solution. 
% }

%为了进一步测试
\section{Methods}

\begin{figure*}[h]
\centering
\includegraphics[width=0.92\textwidth,height=0.36\linewidth]{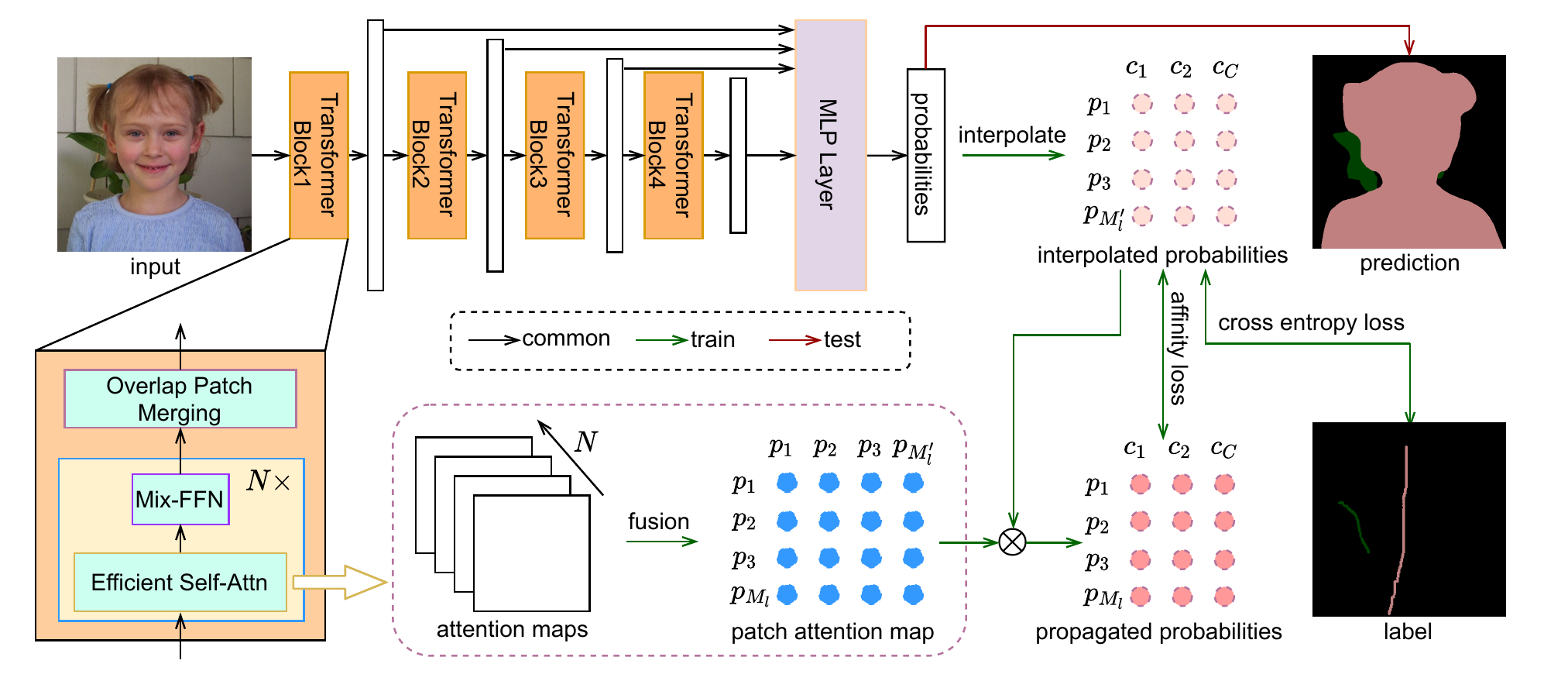} %[width=8\linewidth]
\caption{An overview of the proposed Sparsely Annotated Segmentation Transformer(SasFormer)} \label{fig:framework} %  % 大图名称 clip, trim=0.3cm 0.4cm 0.3cm 6.2cm,
  %图片引用标记
\end{figure*}
% This section introduces SASFormer, our simple but efficient segmentation framework. We first provides an overview in Sec~\ref{sec:overview}. Generation of patch attention map is then introduced in Sec~\ref{sec:pamg}. Finally, analysis of affinity loss is performed in Sec~\ref{sec:afl}.
%In the remainder of this section, we first detail the ...
%首先介绍backbone以及网络的训练和测试过程
%然后介绍patch-level affinity map 如何产生， 
%接着介绍pseudo labels如何产生
%最后是affinity loss
\subsection{Overview}\label{sec:overview}
% We introduce Segformer\cite{xie2021segformer} as our backbone, which is tailored for dense prediction.
As depicted in Fig~\ref{fig:framework}, given an image $I \in \mathbb{R}^{W\times H \times 3}$, we partition it into $p\times p$ resolution patches. These patches are fed into hierarchical transformer encoder to provide multilevel characteristics. We pass multi-level features into the decoder block in order to obtain segmentation probabilities $\mathbf{P}$ with resolution $M\times C$, where $M$ is the patch number and $C$ is the number of categories. 
% In training stage, on the one hand, we generate multi-level patch attention maps with transformer layers in encoder blocks. 
% On the other hand, segmentation probabilities are interpolated into various resolutions which are equal to width of patch attention maps.
On the one hand, we build multi-level patch attention maps with transformer layers in encoder blocks during the training phase.
In addition, segmentation probabilities are interpolated into resolutions corresponding to the width of patch attention maps.
Patch attention maps are then multiplied by interpolated segmentation probabilities, yielding propagated segmentation probabilities $\mathbf{Y}$. 
For labeled regions of interest, we perform standard cross entropy loss function $L_{seg}$ to supervise segmentation probabilities with ground truth labels. For unlabeled regions, we introduce an affinity loss function $L_{aff}$ to match segmentation probabilities to propagated segmentation probabilities.
During testing, segmentation probabilities are immediately assigned to the class with the maximum probability at each pixel, resulting in the final segmentation map. The overall loss function is defined as follows:
%to mining distance between patch tokens who have high 
\begin{equation}\label{eq1}
L = L_{seg} + \alpha *L_{aff}
\end{equation}

\subsection{Patch Attention Map Generation}\label{sec:pamg}
We introduce Segformer~\cite{xie2021segformer} as our backbone in order to describe our method more conveniently. More results of different backbones can be referred to Sec~\ref{sec:ablation}. Hierarchical transformer encoder contains $L=4$ encoder blocks. Each encoder block contains $N_l$ consecutive transformer layers. We get the patch attention map with the efficient self-attention of transformer layer, which is defined as follows:
\begin{equation}
    \mathbf{A}_{l,n} = softmax(\frac{\mathbf{Q}_{l,n} \mathbf{K}_{l,n}^{\mathbf{T}}}{\sqrt{D}})
\end{equation}
where $\mathbf{Q}_{l,n}\in \mathbb{R}^{M_l \times D}$ and $\mathbf{K}_{l,n}\in \mathbb{R}^{M'_l \times D}$ are the query and key representations of $n$-th transformer layer in $l$-th encoder block, respectively. 
$M_l \in \{\frac{W*H}{4}, \frac{W*H}{8}, \frac{W*H}{16}, \frac{W*H}{32}\}$ denotes resolution size in different encoder blocks. 
$M'_l$ denotes resolution size from $M_l$ after sequence reduction process\cite{xie2021segformer} to be more efficient.
$D$ indicates dimension of patch embeddings. $\mathbf{T}$ is the transpose operator. 

Efficient attention map $\mathbf{A}_{l,n} \in \mathbb{R}^{M_l \times M'_l}$ records dependency of each patch token of $n$-th transformer layer in $l$-th encoder block. 
% \suh{Delete: $\mathbf{A}^{i,j}_{l,n}$ means the amount of $j$-th input token embedding contributes to $i$-th output token.}
We aggregate these attention maps over transformer layers in each decoder block as the following:
\begin{equation}
    \mathbf{A}_{l} = \frac{1}{N} \sum_{n=1}^{N}\mathbf{A}_{l,n}
\end{equation}
where $\mathbf{A}_{l}\in \mathbb{R}^{M_l \times M'_l}$ means patch attention map in $l$-th encoder block. 
%%%%%% TBD by chenglc

To better comprehend what information patch attention maps capture at various levels, we randomly draw a point with a red pentagram in the inputs and analyze the dependency between the reference point and other pixels on patch attention maps of various encoder blocks, as depicted in the baseline of Fig~\ref{fig:atten}.
$\mathbf{A}^{i,j}_{l,n}$ means the amount of $j$-th input token embedding contributes to $i$-th output token of $n$-th transformer layer in $l$-th encoder block. Likewise, $\mathbf{A}^{i}_{l}$ means the aggregated amount of each input token embedding contributes to $i$-th output token in $l$-th encoder block. We construct $\mathbf{A}^{r}_{1}$, $\mathbf{A}^{r}_{2}$, $\mathbf{A}^{r}_{3}$ and $\mathbf{A}^{r}_{4}$ to illustrate dependency between the reference point and each pixel in the first, second, third, forth encoder block, respectively.

It can be seen that $\mathbf{A}_{1}$ and $\mathbf{A}_{2}$ prefer to assign larger weights to patches that have the same color and texture as the reference point. For instance, $\mathbf{A}^r_{1}$ and $\mathbf{A}^r_{2}$ of the first input emphasize the white color of the bird, while the second input highlights the dark brown color of the material. $\mathbf{A}_{3}$ and $\mathbf{A}_{4}$ are more concerned with semantic consistency, although $\mathbf{A}_{4}$ is more class-specific. Patch attention maps establish correlations between pixels in color space and high-level characteristics, facilitating the propagation of segmentation information from labeled to unlabeled pixels.
% To better comprehend what information patch attention maps capture in various levels, we randomly draw a point with red pentagram in the inputs and analyze the dependency between reference point and other pixels on patch attention maps of different encoder blocks, as shown in baseline of Fig~\ref{fig:atten}. It can be seen that $\mathbf{A}_{1}$ and $\mathbf{A}_{2}$ tend to deliver higher weights to patches that are similar to the reference point in terms of color and texture. For example, $\mathbf{A}_{1}$ and $\mathbf{A}_{2}$ of first input highlights white color of bird, while second input highlights deep brown color of cloth. $\mathbf{A}_{3}$ and $\mathbf{A}_{4}$ both pay more attention on semantic consistency, while $\mathbf{A}_{4}$ is more class-specific. 
% Patch attention maps provides correlations between pixels from color space to high-level features, which facilitate the propagation of segmentation information from labeled pixels to unlabeled pixels.
% patch attention maps提供了像素之间从颜色空间到高危特征之间的关联性，这些关联性有利于分割信息从labeled pixels 到 unlabeled pixels的传播。但是，patch attention maps 看起来有噪声，。。。
However, transformer tends to distract attention from targets to background area\cite{su2022re}, which can be clearly seen in those of baseline($\mathbf{A}_{2}$) and baseline($\mathbf{A}_{3}$).
%However, patch attention maps appear \cite{su2022re}, especially those of $\mathbf{A}_{2}$ and $\mathbf{A}_{3}$ in baseline. 
% We believe that distractions from irrelevant background partially reflect the limited ability of attention maps to capture dependency, which affects the transmission of information between pixels.
We hypothesize that the inability of attention maps to capture dependence, which impacts the transfer of information across pixels, is partly manifested in distractions from irrelevant background.
%We believe that noise of attention maps partially reflects the limited ability of attention maps to capture dependency, which affects transmission of information between pixels.
%the performance of the network.

\subsection{Affinity Loss Function}\label{sec:afl}
\begin{figure*}[h]
\centering
\includegraphics[width=0.92\textwidth,height=0.36\linewidth]{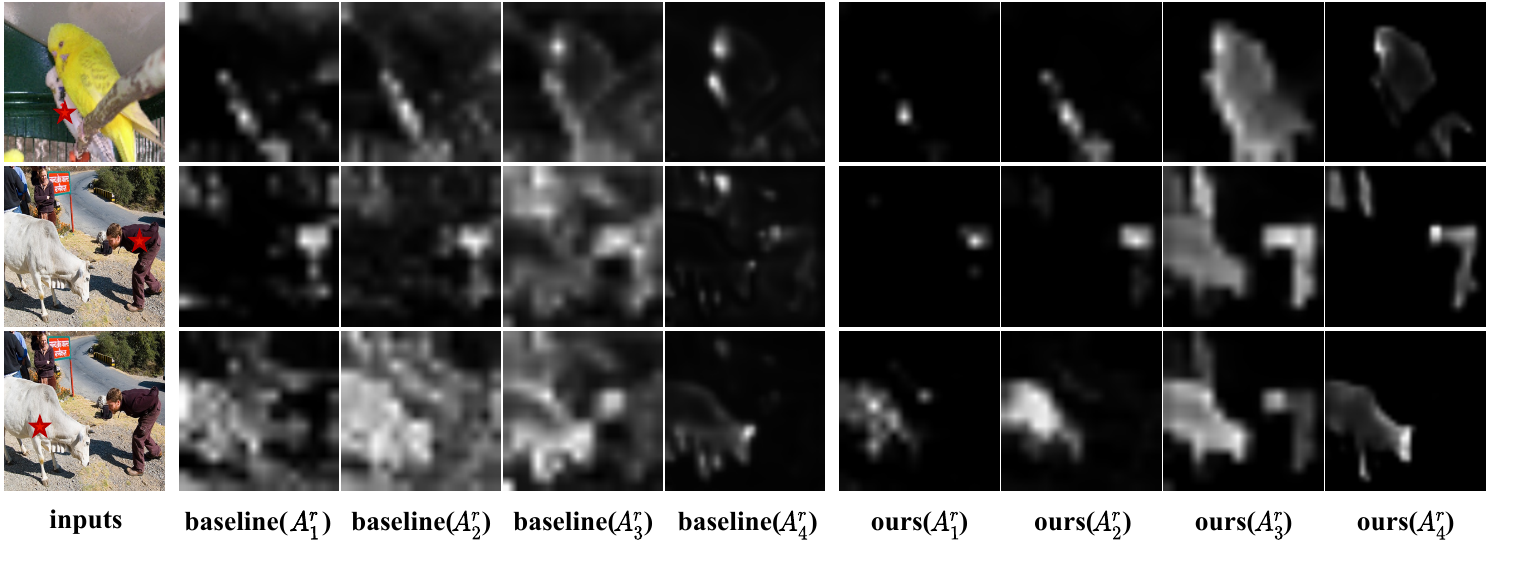} %[width=8\linewidth]
\vspace{-0.5cm}
\caption{Visualization of inputs with reference points (red pentagram) and patch attention maps at various levels. The baseline indicates the Segformer~\cite{xie2021segformer} approach. $\mathbf{A}^{r}_{1}$, $\mathbf{A}^{r}_{2}$, $\mathbf{A}^{r}_{3}$ and $\mathbf{A}^{r}_{4}$ illustrate the dependency between the reference point and each pixel of patch attention map in the first, second, third and forth encoder block, respectively.} \label{fig:atten} %  % 大图名称 clip, trim=0.3cm 0.4cm 0.3cm 6.2cm,

  %图片引用标记
\vspace{-0.5cm}
\end{figure*}
%Therefore, we introduce the affinity loss function to provide the guidance for the attention map.
For each encoder block, we interpolate segmentation probabilities $\mathbf{P}$ into the resolution of $M'_l$ and $M_l$, yielding $\mathbf{P}_{l}^{'}$ and $\mathbf{P}_{l}$ respectively. 
Then we transfer segmentation predictions from labeled pixels to unlabeled pixels by  multiplying patch attention map $\mathbf{A}_{l}$ and interpolated segmentation probabilities $\mathbf{P}_{l}^{'}$, which is described as follows:
\begin{equation}
\mathbf{Y}_{l} = \mathbf{A}_{l} \otimes \mathbf{P}_{l}^{'}
\end{equation}
where $\mathbf{Y}_{l}$ is propagated segmentation probabilities, $\otimes$ denotes matrix multiplication.

$\mathbf{Y}_{l}$ and $\mathbf{P}_{l}$ are normalized along the category dimension, which is defined as:
\begin{equation}\label{affinity_loss}
%\mathbf{Y}_{l}^{*} = softmax(\mathbf{Y}_{l}) 
\mathbf{Y}^{*(m_l,c)}_{l} = \frac{\exp(\mathbf{Y}^{(m_l,c)}_{l})}{\sum_{c=1}^C {\exp(\mathbf{Y}^{(m_l,c)}_{l})}} \qquad 1\le m_l\le M_l
\end{equation}
\begin{equation}\label{affinity_loss}
%\mathbf{P}_{l}^{*} = softmax(\mathbf{P}_{l}) 
\mathbf{P}^{*(m_l,c)}_{l} = \frac{\exp(\mathbf{P}^{(m_l,c)}_{l})}{\sum_{c=1}^C {\exp(\mathbf{P}^{(m_l,c)}_{l})}} \qquad 1\le m_l\le M_l
\end{equation}
where $m_l$ and $c$ denote patch index and class index. $\mathbf{Y}^{(m_l,c)}_{l}$ and $\mathbf{P}^{(m_l,c)}_{l}$ mean the position of row $m_l$, column $c$ of matrix $\mathbf{Y}_l$ and $\mathbf{P}_l$. 

The affinity loss $L_{aff}$ attempts to maximize the similarity between propagated probabilities and the interpolated segmentation probabilities over all encoder blocks. Formally, the affinity loss is defined as:
\begin{equation}\label{affinity_loss}
L_{aff} = \frac{1}{L} \sum_{l=1}^{L} {\Vert \mathbf{Y}_{l}^{*} - \mathbf{P}_{l}^{*} \Vert_1}
\end{equation}
where $\Vert \Vert_1$ and L denote L1 regularization term and encoder block number, respectively. 

In the process of segmentation training, we obtain additional bonus information, as illustrated in Fig~\ref{fig:atten}. Under the constraint of consistency between propagated and interpolated segmentation probabilities, pair-wise dependencies between patch tokens with different labels on patch attention maps are simultaneously deteriorated, while pair-wise dependencies between patch tokens with the same label are enhanced. By comparing to the baseline in Fig~\ref{fig:atten}, it is evident that the background noise in the patch attention map has been suppressed in our proposed SASFormer. 
% Through the construction of affinity loss function, we obtain more potential information of unlabeled pixels in the process of segmentation training. At the same time, under the constraint of consistency between propagated and interpolated segmentation probabilities, pair-wise dependencies between patch tokens with different labels on patch attention maps are weakened, whereas pair-wise dependencies between patch tokens with the same label are strengthened. By comparing baseline and our SASFormer in Fig~\ref{fig:atten}, it is obvious that background noise in patch attention map is restrained. The majority of patch attention map is made up of weights whose labels are the same as reference points.
%With the guidance of affinity loss, pair-wise dependencies between patch tokens with different labels are weakened, whereas pair-wise dependencies between patch tokens with the same label are strengthened. 
% 伪标签的监督在于： 引导网络的attention map

\section{Experiments}
% 我们的方法使用**lr。。。。。。。
% 往期的不同方法都使用了不同的网络结构，不能很好验证我们方法的有效性。
% 因此，除了往期方法的比对，我们还基于相同网络结构去比对baseline,TEL 和我们方法的效果。
%The baseline refers to Segformer-B4 trained with cross-entropy loss only. 
%We report semantic segmentation performance using mean Intersection over Union (mIoU).
\subsection{Datasets and Settings}
The point-wise annotation~\cite{bearman2016s} and scribble-wise annotation~\cite{lin2016scribblesup} of Pascal VOC 2012 dataset~\cite{everingham2010pascal} are utilized for point-supervised and scribble-supervised settings, respectively.
%As for the block-supervised setting, block-wise annotations\cite{liang2022tree} on Cityscapes\cite{cordts2016cityscapes} and ADE20k\cite{zhou2017scene} datasets are employed.
% \begin{table}[t]
% \begin{center}
% \caption{Experimental reults for point-wise annotations on Pascal VOC 2012 validation set} \label{tab:point}
% \begin{tabular}{|c|c|c|c|c|}
%   \hline
%   \multirow {2}{*}{Methods} & \multirow{2}{*}{Backbone} & \makebox[0.05\textwidth]{multi-} & \makebox[0.05\textwidth]{extra} & \multirow{2}{*}{mIoU} \\
%     ~ & ~ & \makebox[0.03\textwidth]{stage} & \makebox[0.03\textwidth]{data} & ~ \\
%   \hline
%   What's the point\cite{bearman2016s} & VGG16 & - & - & 43.4 \\
%   DenseCRF Loss\cite{tang2018regularized} & ResNet101  & $\checkmark$ & - & 57.0\\
%   A2GNN\cite{zhang2021affinity} & ResNet101  & $\checkmark$ & - & 66.8 \\ 
%   Seminar\cite{chen2021seminar} & ResNet101 & $\checkmark$ & -  & 72.5 \\
%   SPML\cite{ke2021universal} & ResNet101  & - & $\checkmark$ & 73.6 \\ 
%   TEL\cite{liang2022tree} & ResNet101  & - & - & 68.4 \\ 
%   \hline
%   Baseline* & Segformer & - & - & 64.2\\
%   TEL*\cite{liang2022tree} & Segformer  & - & - & 69.33\\
%   Sasformer & Segformer  & - & - & 72.76\\
%   \hline
% \end{tabular}
% \end{center}
% \end{table}
\begin{table}[htp]
\begin{center}
\caption{Experimental reults for \textbf{point-wise} annotations on Pascal VOC 2012 validation set.} \label{tab:point}
\begin{tabular}{|c|c|c|c|}
  \hline
  \multirow {2}{*}{Methods} & \multirow{2}{*}{Backbone} & \makebox[0.05\textwidth]{multi-} & \multirow{2}{*}{mIoU(\%)} \\
    ~ & ~ & \makebox[0.03\textwidth]{stage} & ~ \\
  \hline
  DenseCRF Loss~\cite{tang2018regularized} & ResNet101  & $\checkmark$ & 57.00\\
  A2GNN~\cite{zhang2021affinity} & ResNet101  & $\checkmark$ & 66.80 \\ 
  Seminar~\cite{chen2021seminar} & ResNet101 & $\checkmark$ & 72.51 \\
  \hline
  What's the point~\cite{bearman2016s} & VGG16 & - & 43.40 \\
  %SPML\cite{ke2021universal} & ResNet101  & - & 73.6 \\ 
  TEL~\cite{liang2022tree} & ResNet101  & - & 68.40 \\ 
  \hline
  Baseline* & Segformer & - & 64.21\\
  TEL*~\cite{liang2022tree} & Segformer  & - & 69.33\\
  \textbf{Sasformer (Ours)} & Segformer  & - & 73.13\\
  \hline
\end{tabular}
\end{center}
\vspace{-0.5cm}
\end{table}
% \vspace{-0.5cm}
% \subsection{Implementation Details}
We adopt the segformer with pre-trained weights on ImageNet as our backbone for experiments.
Our model is trained for 80k training epochs with initial learning rate 0.001, batch size 16 and input resolution 512x512. The SGD optimizer is used with momentum 0.9 and weight decay 1e-4.
Random horizontal flip, random brightness in [-10, 10], random resize in [0.5, 2.0] and random crop are all employed in data augmentation. In practice, we set $\alpha$ in Eq.\ref{eq1} 1.2 for scribble setting and 0.2 for point setting. All experiments are conducted on PyTorch with four 32GB V100 GPUs. * indicates that we reproduce the results using the provided codes, unless otherwise specified.

\subsection{Comparison with State-of-the-art Methods}
\begin{table}[!htb]
\begin{center}
\caption{Experimental reults for \textbf{scribble-wise} annotations on Pascal VOC 2012 validation set.} \label{tab:scribble}
\begin{tabular}{|c|c|c|c|}
  \hline
  \multirow {2}{*}{Methods} & \multirow{2}{*}{Backbone} & \makebox[0.05\textwidth]{multi-} & \multirow{2}{*}{mIoU (\%)} \\
    ~ & ~ & \makebox[0.03\textwidth]{stage} & ~ \\
  \hline
  ScribbleSup~\cite{lin2016scribblesup} & VGG16 & $\checkmark$ & 63.10 \\
  %NormCut Loss\cite{tang2018normalized} & ResNet101 & $\checkmark$ & 74.5\\
  DenseCRF Loss~\cite{tang2018regularized} & ResNet101 & $\checkmark$ & 75.00 \\ 
  GridCRF Loss~\cite{marin2019beyond} & ResNet101 & $\checkmark$ & 72.80 \\
  URSS~\cite{pan2021scribble} & ResNet101 & $\checkmark$ & 76.10 \\ 
  Seminar~\cite{chen2021seminar} & ResNet101 & $\checkmark$ & 76.20 \\
  A2GNN~\cite{zhang2021affinity} & ResNet101 & $\checkmark$ & 76.20 \\ 
  \hline
  BPG~\cite{wang2019boundary} & ResNet101 & - & 76.00 \\
  SPML~\cite{ke2021universal} & ResNet101 & - & 76.10 \\ 
  PSI~\cite{xu2021scribble} & ResNet101 & - & 74.90 \\ 
  TEL~\cite{liang2022tree} & ResNet101 & - & 77.30 \\ 
  \hline
  Baseline* & Segformer & - & 73.37\\
  TEL*~\cite{liang2022tree} & Segformer & - & 78.53\\
  \textbf{Sasformer (Ours)} & Segformer & - & 79.49 \\
  \hline
\end{tabular}
\end{center}
\vspace{-0.5cm}
\end{table}
% \vspace{-0.5cm}
\begin{figure*}[htp]
\centering
\includegraphics[width=0.92\textwidth,height=0.46\textwidth]{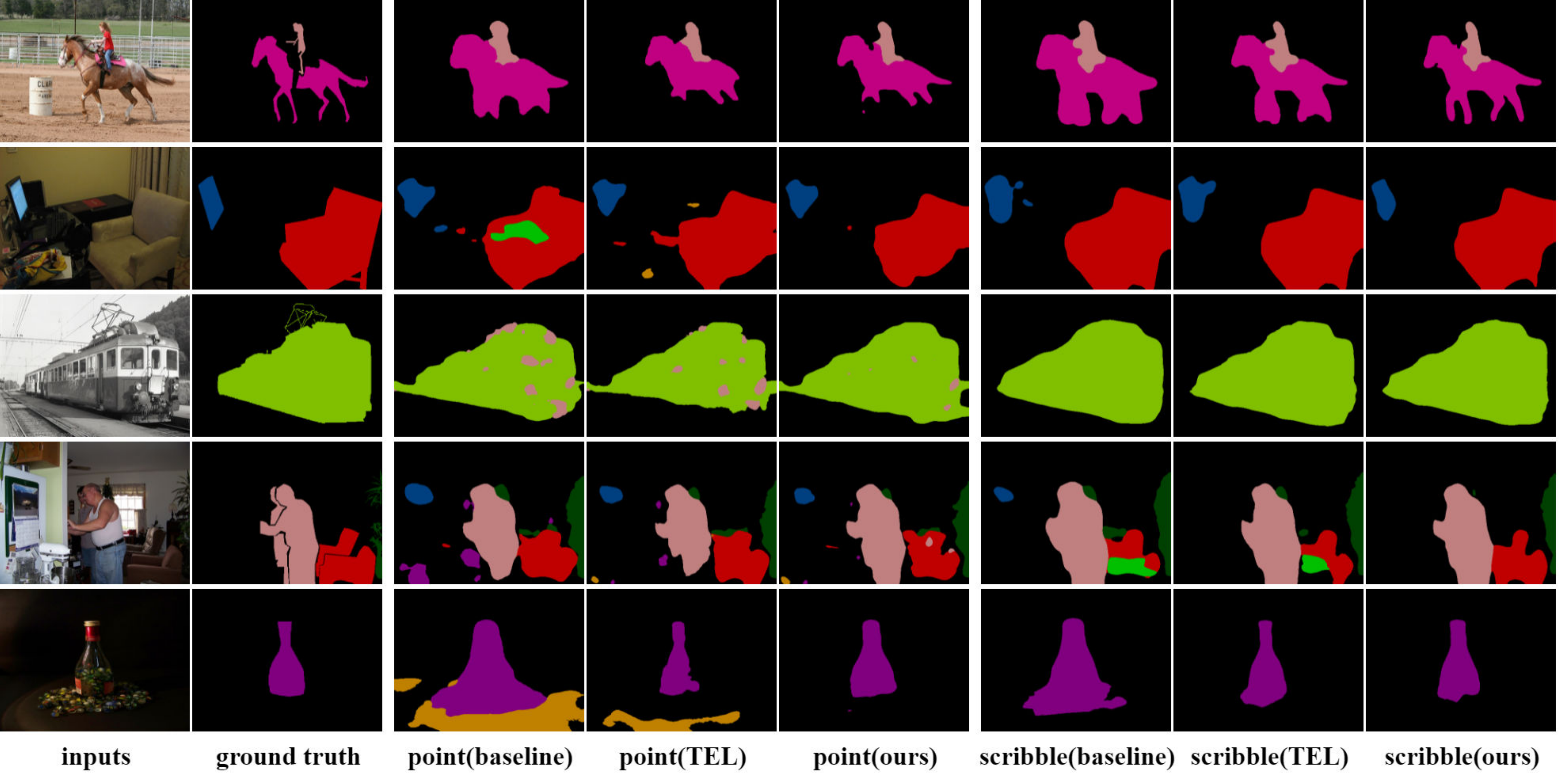} %[width=8\linewidth]
%[width=0.92\textwidth,height=0.36\linewidth]
\caption{Segmentation results on Pascal VOC 2012 dataset. The point and scribble indicate the point and scribble supervision. The baseline and TEL indicate the Segformer~\cite{xie2021segformer} and TEL~\cite{liang2022tree} approach, respectively. } \label{fig:performance} %  % 大图名称 clip, trim=0.3cm 0.4cm 0.3cm 6.2cm,
  %图片引用标记
  \vspace{-0.5cm}
\end{figure*}

Tab~\ref{tab:point} and Tab~\ref{tab:scribble} present the experimental results on the Pascal VOC 2012 for point-wise and scribble-wise annotations, respectively. 
In Tab~\ref{tab:point} for point-wise annotations, the baseline based on segformer-B4 employs the cross-entropy loss function only can reach 64.2\% mIoU. TEL~\cite{liang2022tree} improves mIoU performance by 5.13\%. Our proposed SASFormer outperforms TEL by 4.43\% with the same backbone, yielding a mIoU of 72.78\%. As for scribble-wise annotations in Tab~\ref{tab:scribble}, the baseline framework achieves a mIoU of 73.3\%. SASFormer produces the best performance with the mIoU of 79.49\%, surpassing the baseline by 6.19\%. It demonstrates that SASFormer achieves remarkable performance compared to all other state-of-the-art approaches, which adopt a single-stage training strategy without any additional data. The qualitative results for point-wise annotations and scribble-wise annotations showcase that not only does our technique capture object outlines more accurately, but it also minimizes category misidentification, as shown in Fig~\ref{fig:performance}.

\subsection{Ablations}\label{sec:ablation}
\begin{table}[!htb]
\begin{center}
\caption{Effect of patch attention map in different encoder blocks.} \label{tab:multilevel}
% \begin{subtable}
  \begin{tabular}{c|cccc|c}
  \hline
  \hline
   & $L_{aff}^1$ & $L_{aff}^2$ & $L_{aff}^3$ & $L_{aff}^4$  & mIoU (\%) \\
  \hline
  \hline
  exp1 & - & - & - & -  & 73.37 \\
  \hline
  exp2 & $\checkmark$ & - & - & -  & 77.38 \\
  \hline
  exp3 & - & $\checkmark$ & - & -  & 76.97 \\
  \hline
  exp4 & - & - & $\checkmark$ & -  & 75.98 \\
  \hline
  exp5 & - & - & - & $\checkmark$  & 74.10 \\
  \hline
  exp6 & $\checkmark$ & $\checkmark$ & - & -  & 79.06 \\
  \hline
  exp7 & $\checkmark$ & $\checkmark$ & $\checkmark$ & -  & 79.24 \\
  \hline
  exp8 & $\checkmark$ & $\checkmark$ & $\checkmark$ & $\checkmark$  & 79.49 \\
  \hline
  \hline
  \end{tabular}
% \end{subtable}
\end{center}
\vspace{-0.5cm}
\end{table}

\noindent \textbf{Single scale \textit{vs.} Multi scale.} We examine the influence of patch attention maps in various encoder blocks on segmentation performance in Tab~\ref{tab:multilevel}, where $L_{aff}^{l}$ denotes the affinity loss generated by the patch attention map of  the $l$-th encoder block. The optimal segmentation performance is achieved when patch attention maps are incorporated into the generation of affinity loss in different encoder blocks. 
\begin{table}[!htb]
\begin{center}
\caption{The effect of different loss configurations.} \label{tab:loss_forms}
\begin{tabular}{c|c|c|c|c}
  \hline
  Configuration & KL & CE & L2 & L1 \\
  \hline
  mIoU (\%)  & 78.62 & 76.80 & 77.25 & 79.49\\ % TreeEnergyLoss_n_21 segformer_b4_512_3_aff5.0_kl_8w.log
  \hline
\end{tabular}
\end{center}
\vspace{-0.5cm}
\end{table}
% \vspace{-0.5cm}

\noindent \textbf{The impact of distance metrics.} For the similarity between propagated and interpolated segmentation probabilities, in addition to L1 distance in Eq.\ref{affinity_loss}, cross entropy, L2 distance, and KL divergence may also be considered. To evaluate the performance of our SASFormer, we evaluate different loss configurations. As seen in Tab~\ref{tab:loss_forms}, different forms of affinity loss can also improve performance. We choose L1 distance as the final implementation of our SASFormer since it produces the best results (79.49\% mIoU). 
\begin{table}[!htb]
\begin{center}
\caption{The flexibility of our proposed approach.} \label{tab:applicablity}
\begin{tabular}{c|c|l}
  \hline
  \hline
  Backbone & Ours & mIoU (\%) \\
  \hline
  \hline
  \multirow {2}{*}{Segmentor~\cite{strudel2021segmenter}} & w/o & 72.75 \\ 
  
   ~ & w/ & 76.11 (\textcolor[rgb]{1,0,0}{+3.36}) \\
  \hline
  \multirow {2}{*}{SETR~\cite{SETR}} & w/o & 70.47\\ 
  %\hline
  ~ & w/ & 76.04 (\textcolor[rgb]{1,0,0}{+5.57}) \\
  \hline
  \multirow {2}{*}{segformer~\cite{xie2021segformer}} & w/o & 73.37\\ 
  %\hline
  ~ & w/ & 79.49 (\textcolor[rgb]{1,0,0}{+6.12}) \\
  \hline
  \hline
\end{tabular}
\end{center}
\vspace{-0.5cm}
\end{table}
%\vspace{-1cm}

\noindent \textbf{The flexibility among backbones.} To evaluate the flexibility of our technique across a variety of transformer-based segmentation networks, we also present the popular segmentor~\cite{strudel2021segmenter} and SETR~\cite{SETR} in order to compare segmentation performance with and without affinity loss. As shown in Tab~\ref{tab:applicablity}, our affinity loss is robust across a variety of segmentation networks.

% \input{tables/extra_networks}
% \noindent \textbf{The impact of different backbones.}
% We examine the influence of patch attention maps in various encoder blocks on segmentation performance in Tab\ref{tab:multilevel}, where $L_{aff}^{l}$ denotes the affinity loss generated by the patch attention map of  the $l$-th encoder block. The optimal segmentation performance is achieved when patch attention maps are incorporated into the generation of affinity loss in different encoder blocks. 

% \input{tables/loss_formation}
% \noindent \textbf{The impact of distance metrics.}
% For the similarity between propagated and interpolated segmentation probabilities, in addition to L1 distance in Eq.\ref{affinity_loss}, cross entropy, L2 distance, and KL divergence may also be considered. To evaluate the performance of our SASFormer, we evaluate different loss configurations. As seen in Tab~\ref{tab:loss_forms}, different forms of affinity loss can also improve performance. We choose L1 distance as the final implementation of our SASFormer since it produces the best results (79.49\% mIoU). To evaluate the flexibility of our technique across a variety of transformer-based segmentation networks, we also present the popular segmentor~\cite{strudel2021segmenter} and SETR~\cite{SETR} in order to compare segmentation performance with and without affinity loss. As shown in Tab~\ref{tab:applicablity}, our affinity loss is robust across a variety of segmentation networks.

\section{Conclusions}
In this work, we propose an efficient framework, dubbed SASFormer, to deal with the issue of semantic segmentation under sparsely annotated supervision. 
%Existing approaches based on the pseudo-label are always time-consuming and often necessitate a multi-stage training strategy. 
Most of the existing approaches are always time-consuming and often necessitate a multi-stage training strategy.
While the proposed SASFormer adopts an end-to-end scheme that leverages hierarchical patch attention maps. In addition, we also introduce the affinity loss to capture the consistency between the features of correlation results and network predictions. Exhaustive experiments validate the effectiveness of the proposed approach and showcase remarkable performance compared to state-of-the-art approaches. 

%\input{tables/alg_sas}
%Algorithm~\ref{alg:Alg_SAS}
\section*{Acknowledgments}
This work is partially supported by the National Natural Science Foundation of China (Grant No. 62106235), by the Exploratory Research Project of Zhejiang Lab(2022PG0AN01), by the Zhejiang Provincial Natural Science Foundation of China (LQ21F020003).

\bibliographystyle{IEEEbib}
%\bibliography{icme2022template}
\bibliography{egbib}

% 附录
%\newpage

\clearpage
\appendix
\section{supplementary material }

\begin{table*}[hb]
\begin{center}
%\tiny
\vspace{-0cm}
\caption{Experimental results on Cityscapes and ADE20k datasets with different sparsity levels.} \label{tab:block}
%\resizebox{0.99\linewidth}{!}{%
\begin{tabular}{|c|c|c|c|c|c|c|c|}
  \hline
  \multirow{2}{*}{Method} & \multirow{2}{*}{Params} &
        \multicolumn{3}{c|}{Cityscapes} &
        \multicolumn{3}{c|}{ADE20k}\\ 
  \cline{3-8}
        %\hline
        ~ & ~ & 10\% & 20\% & 50\% & 10\% & 20\% & 50\%\\
  \hline
  DenseCRF Loss~\cite{tang2018regularized} & 70.0M & 57.4 & 61.8 & 70.9 & 31.9 & 33.8 & 38.4 \\
  TEL~\cite{liang2022tree} & 70.0M & 61.9 & 66.9 & 72.2 & 33.8 & 35.5 & 40.0 \\
  \hline
  Segformer & 61.4M & 58.4 & 60.5 & 70.4 & 39.8 & 41.4 & 45.3 \\
  %DenseCRF Loss\cite{tang2018regularized} & ~ & ~ & ~ & ~ & ~ & ~\\
  %TEL*\cite{liang2022tree} & ~ & ~ & ~ & ~ & ~ & ~ \\
  Ours & 61.4M & 64.6 & 70.0 & 75.6 & 41.8 & 44.6 & 46.8 \\
  \hline
\end{tabular}%
%}
%\vspace{-0.6cm}
\end{center}
\end{table*}

% \begin{table*}[tp]
%     \centering
%     %\fontsize{6.5}{8}\selectfont
%         \begin{tabular}{c|c|c|c|c|c|c|}
%         \hline 
%         \multirow{2}{*}{Method} &
%         \multicolumn{3}{c|}{Cityscapes} &
%         \multicolumn{3}{c|}{ADE20k}\\ 
%         \cline{2-7}
%         %\hline
%         ~ & 10\% & 20\% & 50\% & 10\% & 20\% & 50\%\\
        
%         \hline
%         \end{tabular}
%         \vspace{2.5mm}
%         \caption{Experimental results on ILSVRC for metrics of Loc.Acc.} \label{tab:sota-imagenet-la}
%         %\end{threeparttable}
% \end{table*}

\begin{figure*}[hb]
\centering
%\includegraphics[width=0.96\linewidth,,height=0.48\linewidth]{images/config.pdf}
%\vspace{-0.5cm}
\includegraphics[width=\textwidth, height=0.6\textwidth]{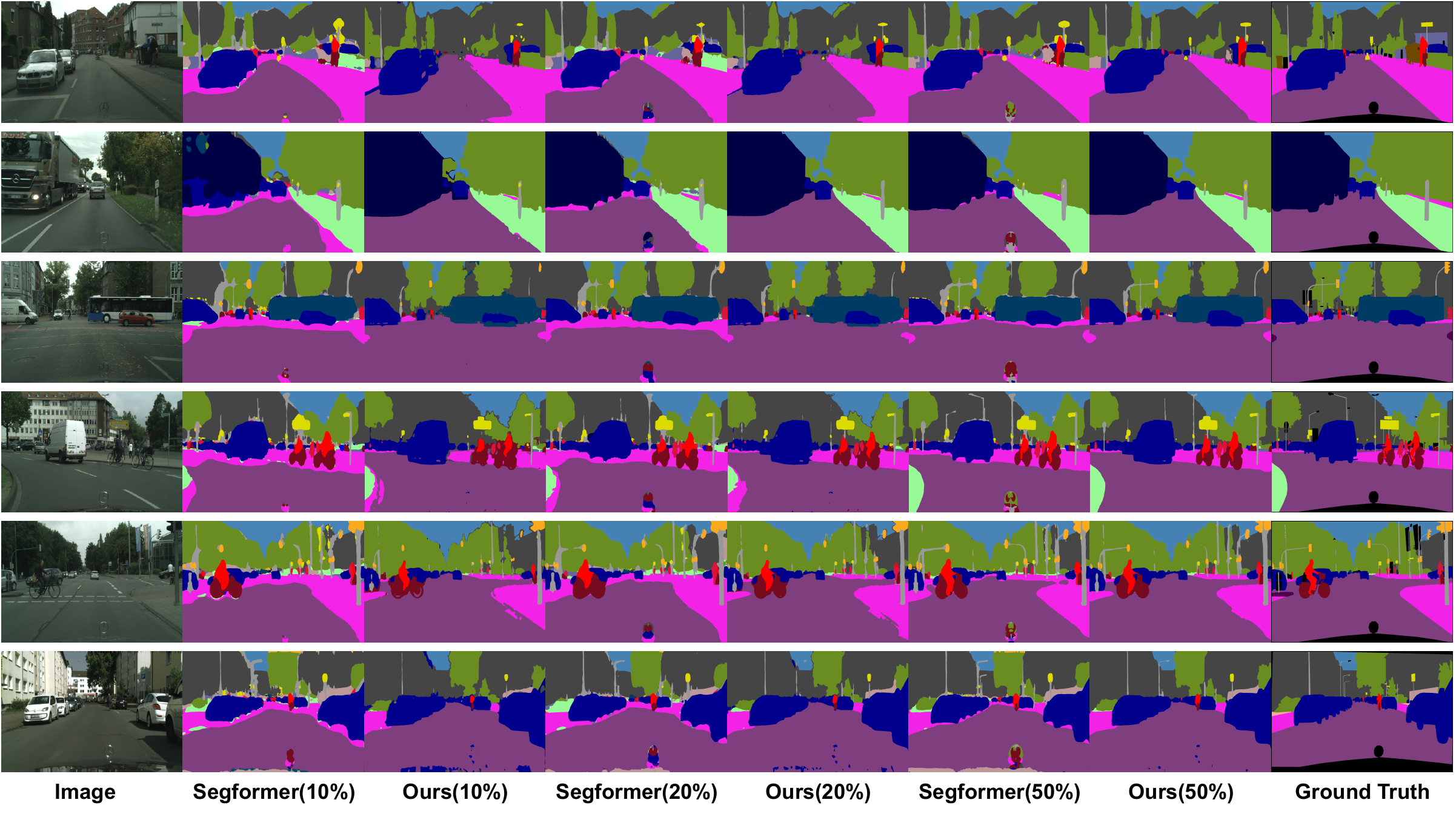} %[width=8\linewidth]
% \caption{Segmentation results on Cityscapes val set. 10\%, 20\%, 50\% are the proportion of full labels, respectively. } \label{fig:citys} %  % 大图名称 clip, trim=0.3cm 0.4cm 0.3cm 6.2cm,
%\vspace{-0.8cm}
\vspace{-0.3cm}
\caption{Cityscapes val set segmentation results at 10\%, 20\%, 50\% label proportions. } 

\label{fig:citys} % 
  %图片引用标记
\end{figure*}

To further demonstrate the effective of our approach, we conduct additional experiments on the Cityscapes and ADE20k datasets, which have segmentation annotations with varying levels of sparsity, including 10\%, 20\%, and 50\% of full labels. As shown in Table~\ref{tab:block}, our SASFormer outperforms state-of-the-art methods at all sparsity levels. Moreover, we include quantitative results on the Cityscapes validation set in  Fig.~\ref{fig:citys}.

\newpage
\begin{figure}[h]
\vspace{0.5cm}
\centering
\includegraphics[width=0.48\textwidth,height=0.5\linewidth]{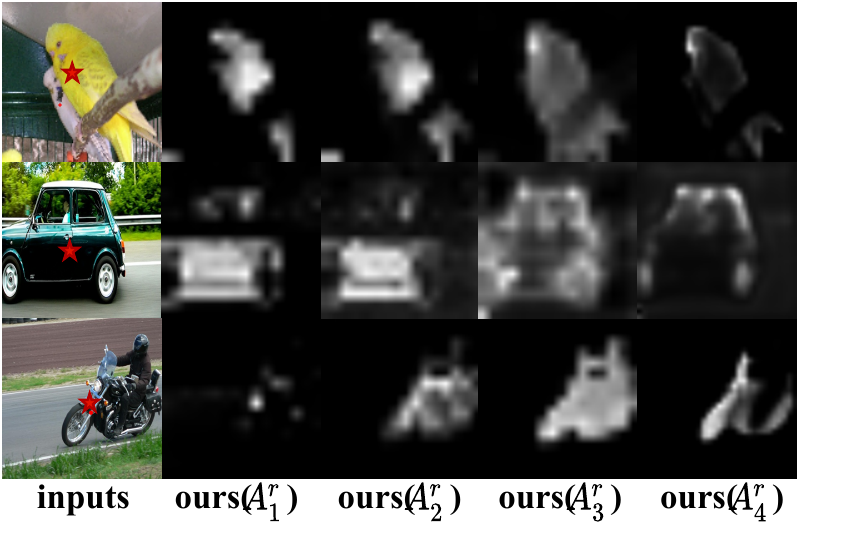} %[width=8\linewidth]
\caption{Visualization of inputs with reference points (red pentagram) and patch attention maps at various levels.} \label{fig:atten_supple} %  % 大图名称 clip, trim=0.3cm 0.4cm 0.3cm 6.2cm,

  %图片引用标记
\vspace{-0.5cm}
\end{figure}

\end{document}